\documentclass[10pt,twocolumn,letterpaper]{article}

\usepackage{cvpr}
\usepackage{times}
\usepackage{epsfig}
\usepackage{graphicx}
\usepackage{amsmath}
\usepackage{amssymb}

\usepackage{makecell}
\usepackage{multirow}
\usepackage[normalem]{ulem}
\usepackage{subcaption}

\usepackage[pagebackref=true,breaklinks=true,letterpaper=true,colorlinks,bookmarks=false]{hyperref}

\cvprfinalcopy 

\def\cvprPaperID{1594} 

\ifcvprfinal\fi
\begin{document}

\title{Sketch-a-Classifier: Sketch-based Photo Classifier Generation}

\author{Conghui Hu$^1$ \quad Da Li$^1$ \quad Yi-Zhe Song$^1$ \quad Tao Xiang$^1$ \quad Timothy M. Hospedales$^{1,2}$\\
$^1$SketchX, Queen Mary University of London \quad \quad 
$^2$The University of Edinburgh \\
{\tt\small \{c.hu, da.li, yizhe.song, t.xiang\}@qmul.ac.uk,
t.hospedales@ed.ac.uk}
}

\definecolor{mypink2}{RGB}{0, 0, 0}
\newcommand{\doublecheck}[1]{\textcolor{black}{{#1}}}
\newcommand{\thnote}[1]{\textcolor{black}{(TH: {#1})}}
\newcommand{\add}[1]{\textcolor{mypink2}{{#1}}}
\newcommand{\cmadd}[1]{\textcolor{black}{{#1}}}
\newcommand{\tabincell}[2]{\begin{tabular}{@{}#1@{}}#2\end{tabular}}
\newcommand{\todo}[1]{\textcolor{red}{({#1})}}
\newcommand{\keypoint}[1]{\vspace{0.1cm}\noindent\textbf{#1}\quad}

\maketitle
\thispagestyle{empty}

\begin{abstract}
Contemporary deep learning techniques have made image recognition a reasonably reliable technology. However training effective photo classifiers typically takes numerous examples which limits image recognition's scalability and applicability to scenarios where images may not be available. This has motivated investigation into zero-shot learning, which addresses the issue via knowledge transfer from other modalities such as text. In this paper we investigate an alternative approach of synthesizing image classifiers: almost directly from a user's imagination, via free-hand sketch. This approach doesn't require the category to be nameable or describable via attributes as per zero-shot learning. We achieve this via training a {model regression} network to map from {free-hand sketch} space to the space of photo classifiers. It turns out that this mapping can be learned in a category-agnostic way, allowing photo classifiers for new categories to be synthesized by user with no need for annotated training photos. {We also demonstrate that this modality of classifier generation can also be used to enhance the granularity of an existing photo classifier, or as a complement to name-based zero-shot learning.}
\vspace{-0.8em}
\end{abstract}

\section{Introduction}
With the maturing of sophisticated deep learning techniques, the conventional image recognition problem has begun to approach a solved problem \cite{szegedy2017inception,deng2009imagenet}. However these great successes depend on large-scale labeled image datasets like ImageNet \cite{deng2009imagenet}. This dependence limits the scalability of the -- otherwise highly successful -- current paradigm. This is because we can't guarantee sufficient annotated examples for all possible concepts -- particularly rare (\eg, rare animals) or emerging (\eg, new man-made objects) categories. 
This limitation has motivated extensive research into zero-shot learning \cite{cvprdavid,fu2016semi,akata2015evaluation,lampert2009learning}, which aims to learn a cross-modal mapping from a domain where categories can be described to the image domain. In this way classifiers can be synthesized given a category description such as attributes \cite{changpinyo2016synthZSL} or word-vectors \cite{socher2013zero}. 

\begin{figure}[t]
\centering
\includegraphics[width=0.9\columnwidth]{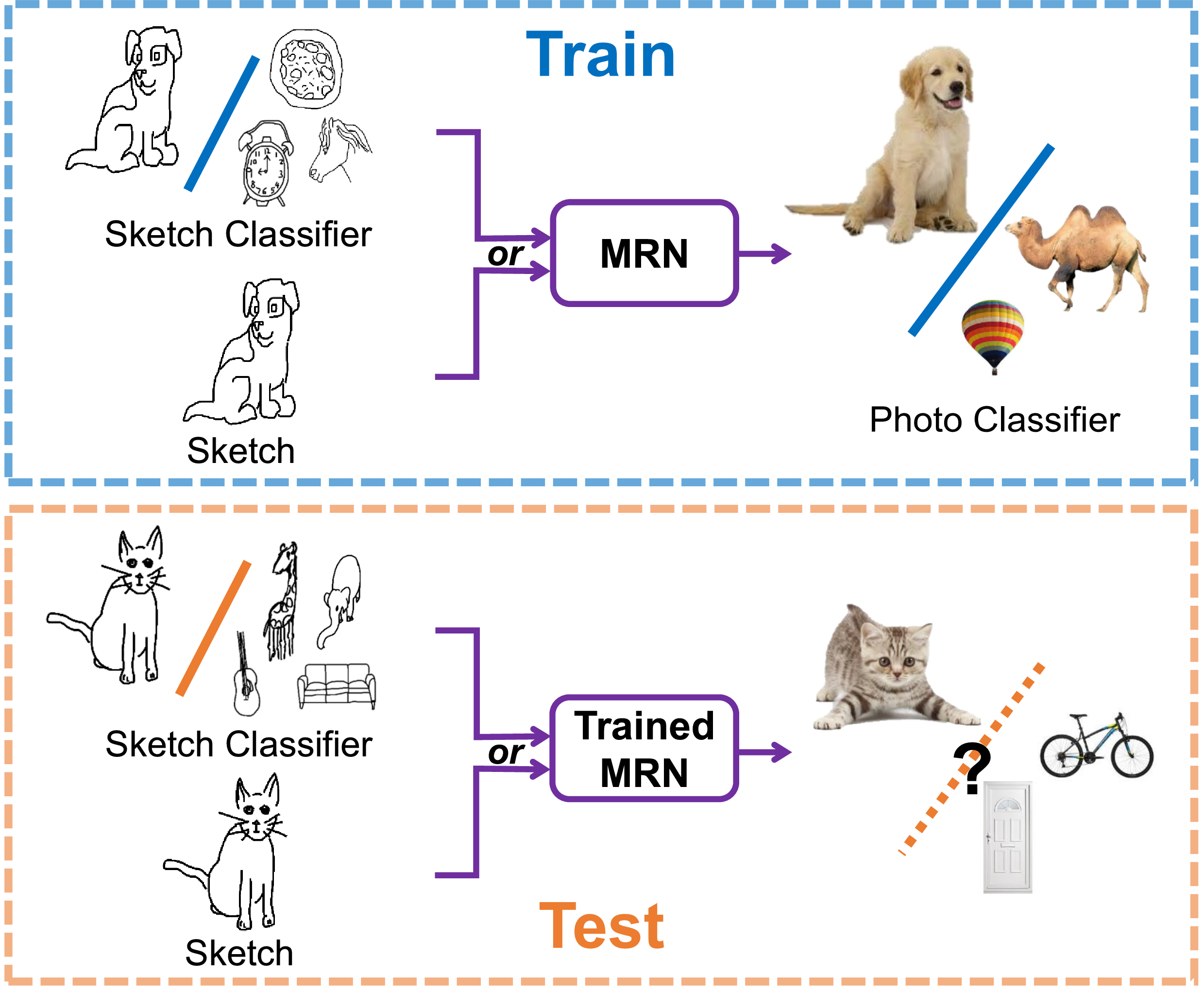}
\caption{Sketch a Classifier: Illustrative Schematic.\label{fig:schematic} MRN: model regression network, which takes a sketch classifier or sketch as input and outputs a photo classifier. For example, during training we have a sketch classifier or sketch of dog, then we train our model regression network to regress to dog photo classifier. At testing time, given a sketch classifier or sketch of an unseen category(\eg, cat), we can use the pre-trained MRN to generate a corresponding photo classifier.}\vspace{-1.em}
\end{figure}

\begin{figure}[t]
\centering
\includegraphics[width=0.9\columnwidth]{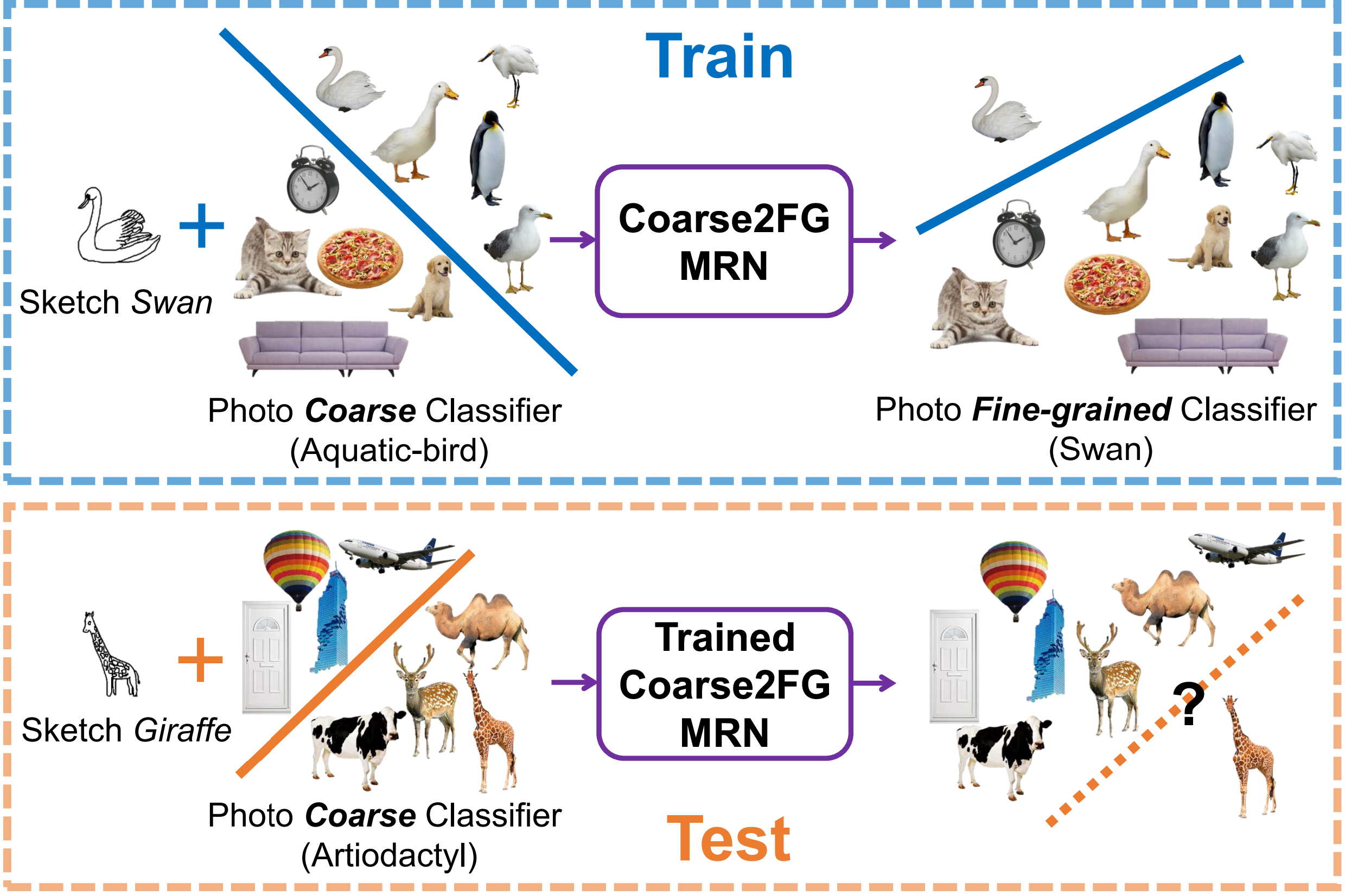}
\caption{Sketch a Fine-grained Classifier: Illustrative Schematic. Coarse2FG: coarse to fine-grained model regression network. For example, during training we have a coarse photo classifier of aquatic bird and a single swan sketch, then we can train our Coarse2FG to regress the fusion of coarse photo aquatic bird classifier and swan sketch to the fine-grained swan photo classifier. At testing time, if a coarse photo classifier of artiodactyl (arbitrary category) is given, we can draw a simple giraffe sketch to generate a giraffe photo classifier. \label{fig:schematic_coarse-fine}}\vspace{-1em}
\end{figure}

The zero-shot learning (ZSL) approach to approach to avoiding per-category data annotation is appealing but has the drawback that it depends on categories being clearly \emph{nameable} or cleanly describable via attributes \cite{akata2015evaluation}. Attributes must be arranged in a pre-determined ontology and so suffer from a different kind of scalability barrier than normal image category annotation \cite{socher2013zero}.  Meanwhile, although more convenient, the efficacy of naming-based approaches that rely on word-vectors depends on having no ambiguity in the category names, and an adequate number of consistent references to the visual concept name in a text corpora to learn meaningful word-vectors. This means that name (word-vector) based ZSL may not be suitable for emerging categories where there is not yet widespread agreement on the name or a large corpora of references to the named concept for word-vector training. It is also unsuited for specialist categories where large text corpora simply may not exist \cite{akata2015evaluation}; or for polysemous concept names (apple computer \vs apple company, financial bank \vs river bank, \etc). 

In this paper we explore an alternative approach to annotation-free generation of photo classifiers that side-steps the above limitations: that of predicting photo classifiers from free-hand sketches. This has the advantage that \textit{neither consistent unambiguous naming, large relevant text corpora, nor structured category description via attributes or parts is necessary}. So long as the user can visually imagine the desired category, they can sketch it and synthesize a classifier. This is related to the task of sketch-based image retrieval (SBIR) \cite{yu2016sketch, sangkloy2016sketchy}, in that both tasks use sketch as input. However SBIR approaches output a list of similar images, while our Sketch-a-Classifier paradigm outputs a classifier that can differentiate different kinds of photos from each other. Moreover, existing SBIR methods either aim to engineer an invariant feature space \cite{hu2013performance,eitz2011sbir} or learn an invariant feature space \cite{yu2016sketch, sangkloy2016sketchy,bui2017compact, deephashingcvpr2017}. The former are not very effective, while the latter depend on being trained on a known set of categories -- they have not been demonstrated to generalize across categories. In contrast, our model-regression approach is designed to learn a category agnostic transformation from sketches to photo classifiers. Given a set of paired sketches and photo classifier examples, we can synthesize classifiers for novel photo categories given an existing sketch classifier, or as little as a single drawing as illustrated in Figure~\ref{fig:schematic}. 

We further show that this paradigm enables various novel extensions including using sketch to define a fine-grained category, and synergistic combination with conventional  zero-shot learning. \textbf{Fine-Grained:} It is often the case that a photo classifier already exists for a broad category due to less difficult annotation, and what is missing is a fine-grained classifier for a rare or emerging sub-category. E.g. bird \vs rare species of bird. 
In this case we can combine the original coarse photo classifier with a sketch of the fine-grained category and produce a fine-grained photo classifier. Again this does not require access to task-relevant text corpora or consistent naming. This process is illustrated in Figure~\ref{fig:schematic_coarse-fine}. \textbf{Enhancing ZSL:} If a category embedding does already exist (\eg, via relevant word-vectors) we show promising results that the category embedding and sketch representation of the category are complementary.

Our contributions are as follows: (i) We introduce the problem of category-agnostic sketch-based classifier synthesis (SBCS).
(ii) We propose an initial model regression based framework for SBCS that can generate classifiers in a zero-shot manner. 
(iii) Several extensions are presented including a fine-grained variant and fusion of SBCS with the standard ZSL paradigm. 
(iv) Promising results are presented on all of these tasks.

\section{Related Work}
\keypoint{Zero-Shot Learning} Our proposed problem is related to zero-shot learning (ZSL) in that it aims to induce photo-domain classifiers. ZSL is now a well studied area which we can only briefly review here. The majority of ZSL approaches exploit category embeddings in the form of word-vectors \cite{frome2013devise,akata2015evaluation,socher2013zero} or attribute-vectors \cite{lampert2009learning,akata2015evaluation}. Common approaches use these category embeddings to learn a cross-domain (image to category embedding) mapping that enables neighbor style matching of images to prototypes of novel categories \cite{socher2013zero,fu2015transductive}; or train a matching function to verify if a given image and category embedding pair match \cite{frome2013devise,akata2015evaluation}. However, as we outlined earlier these approaches have the drawback that they rely on categories being cleanly describable by a pre-established ontology of attributes, or being unambiguously nameable with a large corpus of textual references to the category for word-vector training. In contrast we train  a category-agnostic  sketch$\to$photo model regression network that allows users to synthesize classifiers based solely on their imagination via free-hand sketches. In our experiments, we show that our approach is \emph{complementary} to zero-shot learning in that if we use both category name and freehand sketch illustration as input, we can improve performance compared to either alone.

One ZSL study is related to ours in use of visually abstract (cartoon) person inputs to generate classifiers for photos \cite{antol2014zero}. However, this significantly easier and less general than our task. It only uses cartoon as a manipulation modality to solicit user input. It then uses the annotated pose of persons in the cartoon and photo domain as the representation. This means that: (i) By using a high level pose representation, it does not directly address the whole computer vision problem of sketch and photo interpretation in the respective domains. (ii) This cross-domain mapping of cartoon-person-pose to photo-person-pose is much simpler than the more general mapping between cartoon images and photo images overall. (iii) As a result it is constrained to recognizing photo categories which can be defined by the pose of one or two persons. In contrast, our more general approach does not require any such pose annotation in either modality, and can apply to arbitrary categories. 

\keypoint{Sketch-based image retrieval} {Sketch based image retrieval aims to input a sketch and retrieve photos of the same category as that sketch (category level SBIR \cite{eitz2011sbir,hu2013performance}) or photos corresponding to the specific sketch instance (fine-grained SBIR \cite{yu2016sketch,sangkloy2016sketchy}). Our task is different in that we aim to use sketches to generate photo \emph{classifiers}. And compared to current deep learning-based SBIR methods \cite{yu2016sketch, sangkloy2016sketchy,bui2017compact, deephashingcvpr2017, xu2017cross, song2017sketch} we define a category-agnostic model regression network that can be used to generate classifiers from disjoint categories to those it was trained on.} Existing SBIR models drop rapidly in performance when tested in this `zero-shot' setting on novel categories.

\keypoint{Learning to learn} Existing ZSL methods typically predict prototypes across domains, or train pairwise verification functions. In contrast, our approach is to \emph{predicting the weights} of a classifier \cite{romero2017dietNets,ha2017hypernet} -- specifically the weights of an effective photo classifier for a novel category. This is related to the recently topical area of learning to learn. For example, learned optimizers predict effective weight updates for new tasks during learning \cite{ravi2016optimization,andrychowicz2016learning}. Studies in this area have also addressed few-shot learning. For example  \cite{Hariharan2016LowshotVO} proposed to force the model learning on few samples to have equivalent performance with the model learning on the large scale dataset. Related to ours \cite{wang2016learning} aimed to regress low-shot models onto many-shot models, thus learning a category agnostic `model improvement' transformation that could be used to improve any low-shot model. Our approach is related in that our regression output is a model, \ie, a photo domain classifier.  Beyond  \cite{wang2016learning},  we learn a regressor that is \emph{both} a few $\to$ many-shot  \emph{and} a sketch $\to$ photo domain category agnostic transformation; and we explore using both instances (sketch images), as well as models (sketch classifiers) as input to our model regressor.

\section{Methodology}
The goal of our framework is to produce good photo classifiers, \eg, linear support vector machine (SVM) for binary or multi-way recognition, via  regression networks {given input sketches or other classifier models trained to recognize those sketches.} We consider three kinds of inputs to our regression networks including: SVM model $w$ (binary) or $\mathcal{W}$ (multi-class), image features $\mathcal{\phi}(.)$\footnote{Word-vector can equivalently replace image features, for description simplicity it will not be elaborated here.} or  combination of the coarse-category SVM models and {the fine-grained} sketch image features $\mathcal{C}(w, \mathcal{\phi})$.
Meanwhile, for binary and multi-way recognition purposes, we consider two different regression networks, multi-layer perceptron and convolutional neural networks respectively.
\subsection{Regression Networks}
\keypoint{Model to model regression: Binary} For binary photo recognition problems, we input  SVM parameters trained on sketch domain and predict the parameters of the corresponding SVM for the photo domain. If the regression network is parameterized as $\Theta$, then our model function $\mathcal{F}_{\Theta}(.)$ is to learn the mapping,
\begin{equation}
\label{mlp-eq}
\hat{w}_{p} = \mathcal{F}_{\Theta}(w_{s})
\end{equation}
where, $w_{s}$ and $\hat{w}_{p}$ are the sketch and photo SVM models, \ie, $d+1$ dimensional vector with weight for $d$-dimensional image features and bias, corresponding to the same binary categories. To train the sketch  classifier, $k$ positive sketches and {$j$ negative sketches are randomly selected from the target class and other training categories respectively}.

\keypoint{Model to model regression: Multi-class} For the multi-way photo recognition problem, we regress multi-class sketch-domain SVM models onto multi-class photo-domain SVM models. In this case both input and output are matrices  $\mathcal{W}\in\mathcal{R}^{(d+1)\times c}$ for $c$-way classification and $d$-dimensional features. To deal with these inputs and outputs we design a convolutional network for training. Then the model is to learn
\begin{equation}
\label{conv-eq}
\hat{\mathcal{W}}_{p} = \mathcal{F}_{\Theta}(\mathcal{W}_{s})
\end{equation}
where, $\hat{\mathcal{W}}_{p}$ and $\mathcal{W}_{s}$ are the multi-class  classifier parameters. {Instead of fully connected layers, six convolutional layers with kernel size of ${m\times 1}$ are used. Setting stride as 1, the output size remains the same as the input.} 

\begin{figure*}[t]
\centering
\includegraphics[width=1.0\linewidth]{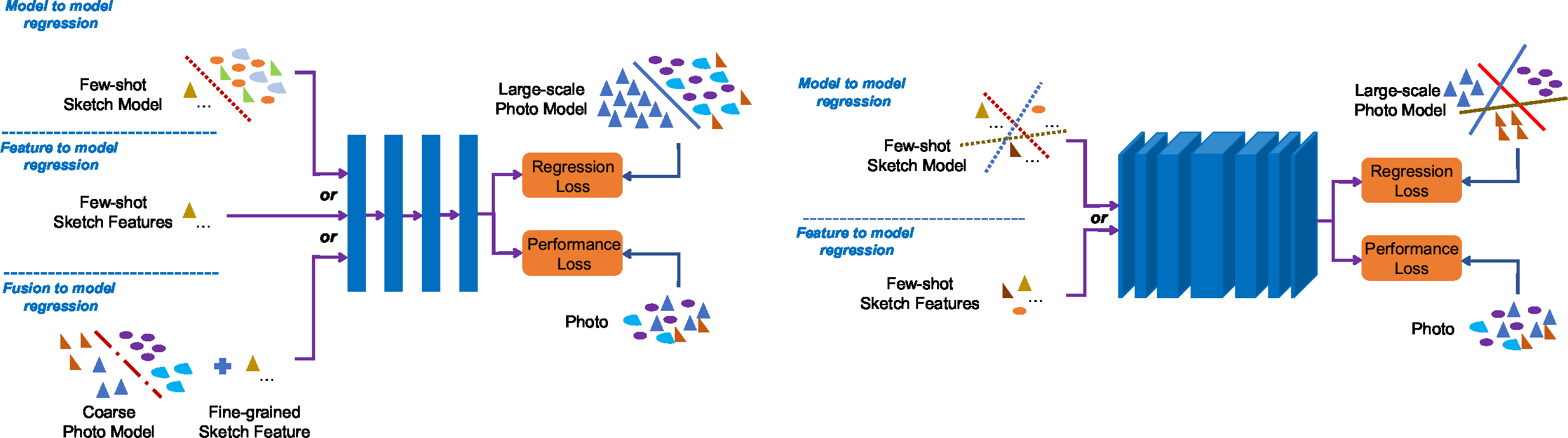}
\caption{Network architecture: Left: binary classifier generation network. Right: multi-class classifier generation network. Better read in zoomed view with color.\label{fig:networkarchitecture}}\vspace{-1em}
\end{figure*}

\keypoint{Feature to model regression} Besides the above model to model regression, we also consider direct feature to model regression. This corresponds to allowing the user to draw a free-hand sketch and directly regressing this sketch onto a photo classifier, rather than training a sketch classifier first before applying the regression. In this case  Eq. \ref{mlp-eq} becomes
\begin{equation}
\hat{w}_{p} = \mathcal{F}_{\Theta}(\sigma^{k}(\phi))
\end{equation}
where, the $\phi(.)$ is the feature extractor. E.g. the \textit{FC7} of \textit{VGG-19} in our experiments to extract the 4096 dimensional features. $k$ is the number of sketch samples used to extract features. $\sigma^{k}(.)$ is the \textit{fusion} function of the $k$ different sketch features generating a $d$ dimensional feature vector, \eg, element-wise \textit{average}. 

For  multi-way classification, we similarly concatenate the feature vectors $\sigma^{k}(.)$ from different $c$ categories to construct a \cmadd{$(d+1)\times c$} matrix as inputs of the convolutional networks. Thus, the Eq. \ref{conv-eq} becomes
\begin{equation}
\hat{\mathcal{W}}_{p} = \mathcal{F}_{\Theta}(\mathcal{C}^{c}(\sigma^{k}(\phi))
\end{equation}
where $\mathcal{C}(.)$ is the concatenation of feature vectors $\sigma^{k}(.)$ from $c$ different categories. 

\keypoint{Fusion to Model Regression for Fine-grained Classifier Tuning} The previous sections described standard sketch feature or sketch model to photo-classifier regression. We also consider the situation where a coarse-grained photo classifier already exists, and this should be combined with a sketch illustrating a fine-grained category to produce a photo classifier for the desired fine-grained category. The intuition is that combining the existing knowledge of the photo-domain super-category with the fine-grained sketch may do better than regressing the fine-grained sketch directly to produce a photo classifier.

In this case the inputs of the regression network become the \textit{fusion} of a prior coarse-grained \emph{photo} SVM model $w^{cg}_p$ and \emph{sketch} image feature $\mathcal{C}(w^{cg}_p, \sigma)$. For example $\mathcal{C}$ may concatenate the $d_p$ dimensional photo SVM weights  $w_p^{cg}$, the $d_s$ dimensional sketch feature $\sigma$ to a $d_p+d_s$ dimensional vector. Then, Eq. \ref{mlp-eq} becomes 
\begin{equation}
\hat{w}^{fg}_{p} = \mathcal{F}_{\Theta}(\mathcal{C}(w^{cg}_{p}, \sigma))
\end{equation}
\subsection{Architecture}
Our model regression networks are illustrated in Figure~\ref{fig:networkarchitecture}. On the left, the binary classification model regression network consists of  fully-connected layers and flexibly fits three different kinds of inputs. The right schematic shows the multi-way classification model regression network. The multi-way model regression network only has convolutional layers. It can handle two different kinds of inputs: few-shot sketch models and few-shot sketch features. For the above two regression networks, \textit{Batch Norm} and Leaky \textit{RELU} layers are applied after each fully-connected and convolutional layer prior to the output layer.

\subsection{Objective Function}
To learn to synthesize effective photo classifiers from sketch, we are inspired by \cite{wang2016learning} to define two kinds of losses: a regression loss and a performance loss.

\keypoint{Regression Loss} This penalizes the $L_2$ distance between the synthesized photo classifier and the ground-truth photo classifier:
\begin{equation}
\mathcal{L}= \lVert \hat{w} - w \rVert_2
\end{equation}
or
\begin{equation}
\mathcal{L}= \lVert \hat{\mathcal{W}} - \mathcal{W} \rVert_F
\end{equation}
where, $w$ and $\mathcal{W}$ are the ground truth photo classifiers for binary and multi-class respectively. 

\keypoint{Performance Loss} Solely requiring that the predicted classifier matches the ground truth may not be sufficient. A small difference in weight values may sometimes have a big difference in classification performance, or vice-versa. Therefore we also define a performance loss to evaluate the practical classification performance of the generated photo classifier on the training photos. For  binary classification, the performance loss is the \textit{hinge} loss
\begin{equation}
\dot{\mathcal{L}}=\max (0, 1-y\cdot{\hat{y}})= \max (0, 1-y\cdot{f_{\hat{w}}(I)})
\end{equation}
where, $I$ is the given photo, $\hat{y}=f_{\hat{w}}(I)$ is the prediction using the generated weights $\hat{w}$ and $y\in\{-1,1\}$ is the ground truth category label for this given photo. 

For multi-way classification,we use \textit{cross-entropy} for the performance loss
\begin{equation}
\dot{\mathcal{L}}= - \sum y\cdot \log (\hat{y}) = - \sum y\cdot \log (f_{\hat{\mathcal{W}}}(I))
\end{equation}
Analogously, $\hat{y}=f_{\hat{\mathcal{W}}}(I) \in \mathcal{R}^{c} $ is the multi-class predication using the generated $\hat{\mathcal{W}}$, where $y \in \mathcal{R}^{c}$ is 1-hot encoded.

\noindent \textbf{Summary}\quad The overall learning objective for the regression network is to input a sketch feature or model and synthesize a photo model that matches the ground-truth model and works as a photo classifier:
\begin{equation}
\underset{\Theta}{\operatorname{argmin}}~~ \alpha \cdot \mathcal{L} + \beta \cdot \dot{\mathcal{L}}
\end{equation}

\section{Experiments}
\subsection{Datasets and Settings}
\keypoint{Datasets}
We use Sketchy dataset \cite{sangkloy2016sketchy} which contains about 75000 sketches sketch and 12500 photos across 125 categories, as well as 56166 additional ImageNet photos of categories in Sketchy. For evaluating category-agnostic model regression, we split Sketchy into training and testing categories (details given in each specific experiment). {We train on sketches from our Sketchy train split and  photos from the corresponding ImageNet categories. We test on sketches and photos from our Sketchy test split.} In this way the photos used in testing are truly novel photos. 


\keypoint{Model Regression Architecture}
For  binary classifier regression, four fully connected layers are used. The number of units for each layer is the same as \cite{wang2016learning}. For multi-class classifier regression the architecture is a six layer fully convolutional (matching size input and output) network with $32,64,128,64, 32, 1$ channel(s) at each respective layer.

\begin{table}
\begin{center}
\scalebox{0.76}{
\begin{tabular}{|c|c|}
\hline
Coarse & Fine-grained Category Names \\
\hline
C1 & airplane, blimp, helicopter, sailboat \\
C2 & scorpion, spider, crab, hermit\_crab, lobster \\
C3 & cannon, knife, pistol, rifle, rocket, sword \\
C4 & rabbit, mouse, squirrel, hedgehog \\
C5 & axe, hammer, racket, saw, scissors, teapot \\
C6 & bench, chair, couch, table, wheelchair \\
C7 & cabin, door, skyscraper, window \\
C8 & armor, hat, shoe, umbrella \\
C9 & bread, hamburger, hotdog, pizza, pretzel \\
C10 & apple, banana, pear, pineapple, strawberry \\
C11 & duck, penguin, seagull, swan, wading\_bird \\
C12 & crocodilian, lizard, sea\_turtle, snake, turtle \\
C13 & camel, cow, deer, giraffe \\
C14 & bear, cat, lion, raccoon, tiger \\
C15 & ant, bee, beetle, butterfly \\
C16 & parrot, owl, chicken, songbird  \\
C17 & horse, rhinoceros, zebra, elephant \\
C18 & fish, ray, shark, dolphin, seal \\
C19 & guitar, harp, piano, saxophone, trumpet, violin\\
C20 & bicycle, car\_(sedan), motorcycle, pickup\_truck, tank \\
\hline
\end{tabular}
}
\caption{\small Coarse/fine-grained grouping for Sketchy dataset categories. We split  Sketchy  into 20 coarse-category groups, where each group has 4 to 6 fine-grained categories.}
\label{grouping result}
\end{center}
\vspace{-1.8em}
\end{table}

\keypoint{Features} For photo features we use the ILSVRC 1000-category pre-trained VGG19 \cite{simonyan2014very} model to extract FC7 layer features. 
The photo model is not well tuned for sketch feature extraction, so we fine-tune the VGG-19 model for sketch recognition on Sketchy dataset (excluding the testing categories), and apply the fine-tuned model for sketch feature extraction. For word-vectors we use the {word2vec} model pre-trained on Google News corpus (3 billion running words) to get one {$300d$} word-vector \cite{mikolov2013distributed} (as per most recent ZSL \cite{socher2013zero,akata2015evaluation} work) for each of 125 categories.

\keypoint{Training Setting: Regression model} Adam optimizer is used in all experiments with initial learning rate of $2\times 10^{-5}$, hyper parameters $\alpha=0.01$, $\beta=1$. The mini batch size for multi-class classifier regression and binary classifier regression are 16 and 64 respectively.

\subsection{Models for Comparison}
To evaluate the efficacy of our proposed method. We consider the following models as baselines for comparison:

\keypoint{Feature-Based}~\\
\textbf{Sketch Nearest Neighbor:} We take the target category sketches, extract deep features and treat these as labeled photos. These are compared directly to the deep features of the photos to classify.\\
\textbf{Sketch Nearest Neighbor + Subspace Alignment:} Vanilla nearest neighbor may not work well due to the domain shift between sketch and photo. Subspace alignment \cite{fernando2013unsupervised} aims to improves cross-domain matching by aligning the subspaces for comparison.\\
\textbf{Triplet Ranking}: These methods \cite{yu2016sketch,sangkloy2016sketchy} use a three branch network and triplet loss to learn a good aligned sketch-photo representation for sketch-photo matching. For fair comparison, we re-train the Sketch Me That Shoe network \cite{yu2016sketch} using the same training categories. 

\keypoint{Model-Based}~\\
\textbf{Sketch-model:} SVM models trained with one or five positive sketches.\\
\textbf{Photo-model:} The upper bound, assuming we have photos to train a few shot model. 

\keypoint{Feature$\to$Model Regression}~\\
\textbf{Word-vector:} This corresponds to the standard regression-based approach to zero-shot learning such as:  \cite{socher2013zero,fu2015transductive}. Such ZSL approaches are not a direct competitor for our approach as we do not rely on word-vectors, but it provides some context for performance.\\
\textbf{Sketch-feature:} Our framework, regressing the features of $k=\{1,5\}$ shot sketches as inputs.\\
\textbf{Photo-feature:} This extends the  few$\to$many shot regression as per \cite{wang2016learning} to the case of using feature (few-shot example) rather than model (few-shot classifier) inputs. It provides an upper bound of how well we could do if we actually have photos of the target categories to recognize.

\keypoint{Model$\to$Model Regression}~\\
\textbf{Sketch-model:}  Our framework, regressing the $k=\{1,5\}$ shot trained sketch classifier to a many-shot photo domain classifier before applying it.\\
\textbf{Photo-model:} The upper bound of photo model based few$\to$many shot regression \cite{wang2016learning}.

\subsection{Results}
\subsubsection{Binary Photo Classifier from Sketch}

In the first experiment we evaluate synthesizing 1-\vs-all photo classifiers based on sketches.

\keypoint{Settings:} Of the 125 Sketchy categories, we use 115 for training and 10 for testing. To train the input and target SVMs, we keep the number of negative examples at 600 for all few-shot and ground-truth models. For training the model regression network, we need multiple few-shot sketch models. We train 500 input SVM models for each category with different regularization parameters from $10^{-2, -1, 0, 1, 2}$ and {different randomly selected positive sketches}. For ground-truth (target) photo model, all ImageNet photos of the target category are taken as positive examples when training one many-shot photo recognition model per category. The same ground-truth photo models are used in the feature and model-based regression input.

\keypoint{Evaluation Metrics} We use average precision of binary photo classification.
Average precision is computed by ranking the the testing set according to the classifier score \doublecheck{and compute the average precision over all recalls.} 
Reported results are averages over the performance of 100 regressed features/models for different choices of input sketches  (except W.V. as there is only one per category). 

\begin{figure}
\hspace{0.8cm}
\includegraphics[width=0.75\columnwidth]{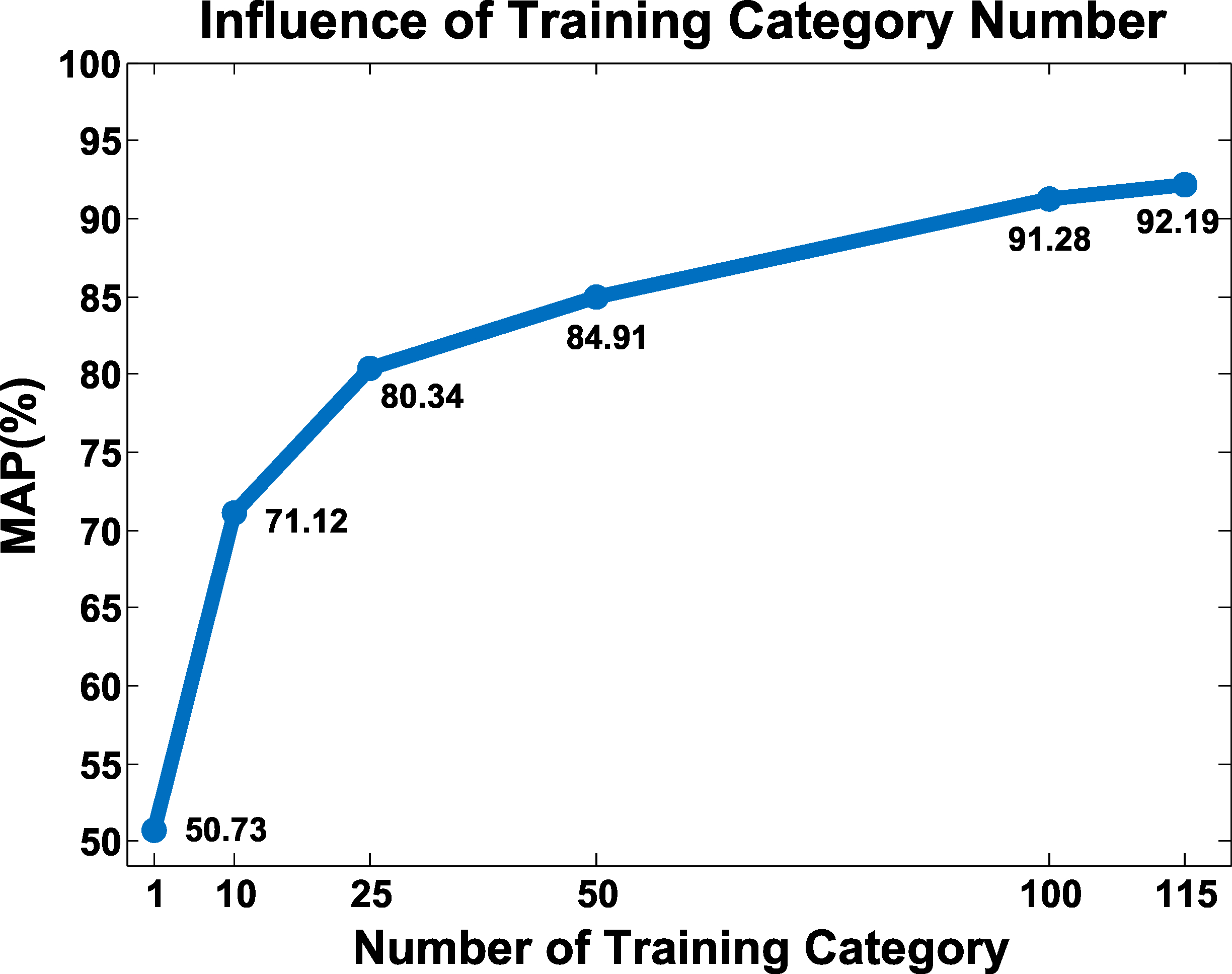}
\caption{Dependence on number of train categories (binary one-shot feature regression)}
\label{influence-sketch-cate-num}
\vspace{-0.8em}
\end{figure}

\begin{table*}[th]
\begin{center}
\vspace{-0.2cm}
\scalebox{0.75}{
\begin{tabular}{c|c|c|c|c|c|c|c|c|c|c|c|c}
\hline
 & \makecell{Classification\\ Method} & Car\_(sedan) &  Pear & Deer & Couch & Duck & Airplane & Cat &  Mouse & Seagull & Knife & 
 \makecell{Binary:\\ mAP$(\%)$} \\
\hline
\multirow{4}{*}{\makecell{\textit{Non} \\Reg.}}
& S.F. NN & 78.27  & 49.64 & 55.67 & 64.97 & 34.24 & 43.63 & 46.60 & 40.32 & 56.24 & 58.91 & 52.85 \\
& S.F. NN+SA \cite{fernando2013unsupervised} & 84.15 & 48.74 & 51.76  & 69.25 & 41.88 & 36.98 & 49.87 & 39.25 & 68.87 & 58.22 & 54.90  \\
& SAN-S \cite{yu2016sketch} & 89.64  & 82.97 & 86.84 & 84.27 & 75.04 & 77.58 & 74.39 & 66.05 & 75.78 & 75.35 & 78.79  \\
\hline
\multirow{4}{*}{\makecell{\textit{Non} \\Reg.}}
& \textit{one}-shot S.M. & 99.24 & 81.77 &  86.46 & 92.31 & 55.98 & 69.28 & 88.45 & 69.11 & 63.35 & 79.69 & 78.56 \\
& \textit{five}-shot S.M. & 99.95 & 94.88 &  97.65 &  97.85 & 79.60 & 95.23 & \textbf{96.21} & 82.57 & 76.12 & 92.06 & 91.21 \\
& \textit{one}-shot P.M.  & 99.98 & 98.06 & 98.66 & 97.49 & 87.69 & 98.96 & 94.20 & 92.91 & 91.82 & 96.39 & 95.62 \\
& \textit{five}-shot P.M. & 100.00 & 99.79 & 99.87 & 99.59 & 99.00 & 99.95 & 98.15 & 98.80 & 99.02 & 99.01 & 99.32 \\
\hline
\multirow{4}{*}{\makecell{\textit{M2M} \\Reg.}}
& \textit{one}-shot S.M. & 97.78 & 93.42 & 92.66 & \textbf{99.51} & 65.10 & 80.44 & 73.13 & 69.66 & 54.06 & 90.16 & 81.59 \\
& \textit{five}-shot S.M. & 99.98 & 94.59 & 93.17 & 98.92 & 80.11 & 96.22 & 85.54 & 88.46 & 67.42 & 94.64 & 89.91 \\
& \textit{one}-shot P.M. & 99.98 & 99.17 &  99.78 & 98.58 & 84.80 & 99.92 & 87.72 & 95.61 & 74.13 & 98.73 & 93.84 \\
& \textit{five}-shot P.M. & 100.00 & 99.64 & 99.93 & 99.72 & 96.55 & 99.90 & 94.02 & 98.27 & 84.90 & 99.00 & 97.19 \\
\hline
\multirow{7}{*}{\makecell{\textit{F2M} \\Reg.}}
& W.V. & 99.91 & \textbf{98.37} & 92.75 & 99.24 & 69.15 & \textbf{99.08} & 80.02 & 83.03 & 86.11 & \textbf{98.38} & 90.60 \\
& \textit{one} S.F. & 97.51 & 95.00 & 96.38 & 95.56 & 93.13 & 93.40 & 89.05 & 87.41  & 80.17 & 94.30 & 92.19 \\
& \textit{five} S.F. & \textbf{100.00} & 96.65 & \textbf{99.85} & 99.38 & \textbf{96.41} & 98.34 & 91.17 & \textbf{92.16} & \textbf{87.05} & 92.54 & \textbf{95.35} \\
& \textit{one} S.F.+W.V. & 98.33 & 95.09 & 97.30 & 97.33 & 95.02 & 94.90 & 88.39 & 85.28 & 81.41 & 92.10 & 92.52 \\
& \textit{one} P.F. & 99.93 & 98.24 & 99.00 & 99.32 & 96.58 & 99.83 & 87.57 & 93.76 & 88.25 & 96.37 & 95.89 \\
& \textit{five} P.F. & 99.99 & 99.67 & 99.86 & 99.81 & 98.37 & 99.94 & 94.86 & 98.00 & 93.01 & 98.47 & 98.20 \\
\hline
\end{tabular}
}
\caption{\small Photo classification accuracy on Sketchy Dataset: Binary. Metrics: mean Average Precision (\%). Non Reg.: no regression used. M2M Reg.: model to model regression. F2M Reg.: feature to model regression. S.M.: sketch model. P.M.: photo model. S.F.: sketch feature. P.F.: photo feature. W.V.: word-vector. Entries based on photos are upper bounds for context. Best non-photo results are in bold.}
\label{category-level binary}
\end{center}
\end{table*}
\keypoint{Results:} The classification results are shown in Table \ref{category-level binary}. From the results we make the following observations: 
(i) Comparing the direct cross-domain application of sketch models to the regressed models (Non Reg. S.M. \vs M2M Reg. S.M.), we see that the regression network significantly improves performance in the 1-shot but not 5-shot case.
(ii) Generally the regression network worked better for sketch feature input than model input (Sketch F2M \vs M2M). It also trains a classifier that is much better than using the raw input sketch feature for NN matching (F2M One S.F. \vs Sketch NN).
(iii) Contrary to \cite{wang2016learning}, we found limited improvement from photo-based few$\to$many-shot model regression.
(iv) Although word-vector-based ZSL is not a direct competitor to sketch-based classifier generation (since it depends on name-ability), it is interesting that the F2M regressed sketch inputs (F2M Reg. Sketch Feature) outperform it. The margin becomes larger if the user spends more effort to provide five rather than one input sketch -- there is no analogy to this in conventional ZSL. 
(v) Our regression network is capable of combining sketch inputs and word-vector inputs in a complementary way: F2M One S.F. + WV outperforms F2M One S.F. and F2M W.V. alone. 
(vi) The triplet ranking approach \cite{yu2016sketch} learns a shared embedding that improves significantly on vanilla NN matching or subspace alignment, but it is still not competitive with the model-regression approaches.
(vii)  If the user is willing to draw more than one sketch to convey some intra-class variability when defining classifier, performance improves for both feature and model input. In particular the classifier generated by regressing 5 sketch features is comparable to using one-shot photo. Thus a `photo is worth five sketches'. 
(viii) The Photo classification accuracy is affected by how easily confusable the test categories are, \eg relatively large number of mis-classifications between the similar 'Duck' and 'Seagull' classes.
(ix) An analysis on the impact of sketch quality on classification accuracy can be found in Table \ref{influence-sketch-quality}. Higher quality sketches produce better photo classifier, but the performance gap is obvious only for the bottom 10 sketches.
(x) As shown in Figure~\ref{influence-sketch-cate-num}, more training categories would improve results  due to having more data for training and increased chance of including a similar category to a given test category. 

\begin{table*}[th]
\begin{center}
\vspace{-0.17cm}
\scalebox{0.80}{
\begin{tabular}{c|c|c|c|c|c|c|c|c|c|c|c}
\hline
 & Car\_(sedan) &  Pear & Deer & Couch & Duck & Airplane & Cat &  Mouse & Seagull & Knife & \makecell{Binary mAP$(\%)$} \\
\hline
Bottom 10 & 85.87  & 89.82 & 84.03 & 91.58 & 93.41 & 73.83 & 82.17 & 84.13 & 81.39 & 89.28 & 85.55 \\
Middle 10 & 96.44 & 94.27 & 98.47  & 96.88 & \textbf{95.59} & 97.93 & 87.93 & 89.10 & 73.06 & 96.60 & 92.63 \\
Top 10 & \textbf{99.95} & \textbf{96.23} & \textbf{99.57} & \textbf{98.22} & 94.72 & \textbf{99.87} & \textbf{93.32} & \textbf{94.50} & \textbf{85.54} & \textbf{97.07} & \textbf{95.90} \\
\hline
\end{tabular}
}
\caption{Influence of sketch quality (binary one-shot feature regression). VGG-19 trained for sketch recognition is used as an indicator of sketch quality.}
\label{influence-sketch-quality}
\end{center}
\vspace{-1.8em}
\end{table*}

\vspace{-0.7em}
\subsubsection{Multi-class Photo Classifier from Sketch}
We next evaluate synthesizing multi-way photo classifiers using our convolutional model regression network.

\keypoint{Settings:} We use the same 10 categories as the binary classifier regression for the test set. To train the model regression, among the 115 training categories we randomly select 10 categories to train 100 10-way multi-class sketch classifier and one 10-way multi-class photo classifier.  All together 500 random groups are created in order to generate classifiers and train this model regressor.  All photos from selected categories are used to train the ground-truth multi-class classification model for both feature and model regression. 

\keypoint{Results:}  
From the results shown in Table \ref{category-level multi}, we can draw the conclusions: (i) Our model to model regression successfully improved the multi-class sketch recognition model for application to photos (M2M Reg. S.M. improves Non Reg. S.M.). However we found that the photo performance was little affected by model regression. (ii) Our feature to model regression outperformed model to model regression (F2M Reg. S.F. \vs M2M Reg. S.M.). (iii) As in the previous experiment, if multiple sketches are available to encode some intra-class variability performance is greatly improved (five S.F. \vs one S.F.). (iii) Overall in this case the sketch-based model regression outperformed the word-vector alone baseline (F2M Reg. S.F. \vs F2M Reg. W.V.). (v) Again comparing to the upper bound that assumes photo availability, we see that model-regression based on five sketches performs comparably to the availability of a single target class photo.
\vspace{-1.8em}
\subsubsection{Coarse to Fine-Grained Photo Classification through Sketching}
\doublecheck{The goal of this task is to transform a coarse-grained photo classifier to a fine-grained classifier using a sketch. This is motivated by the idea of defining a classifier for a new (\eg man made object) or rare (\eg animal) fine-grained category within a known coarse category. In the absence of specific datasets for this we illustrate the concept using a coarse/fine-grained category grouping within Sketchy dataset.} Specifically, we group the categories in Sketchy according to the WordNet structure in ImageNet. This gives us 20 groups (coarse categories) containing 95 sub-categories (fine-grained categories) as illustrated in  Table~\ref{grouping result}.
\begin{table}
\begin{center}
\scalebox{0.80}{
\begin{tabular}{c|c|c}
\hline
& Classification Method & Multi-class: Accuracy$(\%)$ \\
\hline
& Sketch NN & 16.25  \\
& SAN-S \cite{yu2016sketch} & 23.95 \\
\hline
\multirow{2}{*}{\textit{Non} Reg.}
& \textit{five} S.M. & 78.60 \\
& \textit{five} P.M. & 93.30 \\
\hline
\multirow{2}{*}{\textit{M2M} Reg.} &  \textit{five} S.M. & 79.93 \\
&  \textit{five} P.M. & 93.55 \\
\hline
\multirow{5}{*}{\textit{F2M} Reg.} & W.V. & 35.90 \\
& \textit{one} S.F. & 68.16 \\
& \textit{five} S.F. & \textbf{83.01} \\
& \textit{one} P.F. & 84.12 \\
& \textit{five} P.F. & 93.89 \\
\hline
\end{tabular}
}
\caption{\small Photo classification accuracy on Sketchy Dataset: Multi-class. Abbreviations as in Table~\ref{category-level binary}. Best non-photo result is in bold.}
\label{category-level multi}
\end{center}
\vspace{-2em}
\end{table}
\begin{table*}[th]
\begin{center}
\vspace{-0.2cm}
\scalebox{0.75}{
\begin{tabular}{c|c|c|c|c|c|c|c|c|c|c|c|c|c|c}
\hline
Auxiliary input type & CG & helicopter & airplane & blimp & sailboat & giraffe & deer & cow & camel & window & door & skyscraper & cabin & Binary:mAP (\%) \\
\hline
NN + S.F. & $\times$& 15.37 & 5.83 & 9.75 & 25.79 & 8.23 & 7.38 & 6.09 & 5.11 & 22.54 & 11.24 & \textbf{22.23} & 17.52 & 13.09   \\
W.V. & $\times$& 34.28 & 12.61 & 10.23 & \textbf{69.10} & 8.09 & 8.03 & 20.75 & 21.07 & \textbf{33.58} & \textbf{29.64} & 10.32 & 13.87 &  22.63 \\
\textit{one} S.F. & $\times$& 32.78 & 40.16 & 38.23 & 34.79 & 15.01 & 34.52 & 35.72 & 28.95 & 29.82 & 21.25 & 17.85 & 32.98 &  30.16 \\
\textit{five} S.F. & $\times$& \textbf{46.15} & \textbf{58.90} & \textbf{55.02} & 30.42 & \textbf{17.07} & 50.09 & \textbf{39.82} & \textbf{52.90} & 20.39 & 23.09 & 16.72 & 44.82 & \textbf{37.95} \\
one S.F. + W.V. & $\times$& 34.27 & 46.08 & 35.75 & 18.13 & 16.29 & \textbf{53.26} & 35.17 & 29.01 & 25.91 & 25.84 & 14.18 & \textbf{44.98} & 31.57 \\
\textit{one} P.F. & $\times$& 30.58 & 69.67 & 55.88 & 61.29 & 38.00 & 35.36 & 50.51 & 58.09 & 26.60 & 53.32 & 25.11 & 69.67 & 47.84 \\
\textit{five} P.F. & $\times$ & 25.46 & 80.50 & 53.67 & 89.56 & 25.99 & 34.34 & 49.63 & 73.58 & 34.42 & 55.19 & 38.23 & 77.87 & 53.20 \\
\hline
NN + S.F.  & \checkmark & 35.71 & 17.64 & 23.66 & 51.32 & \textbf{41.71} & 32.68 & 23.41 & 17.37 & 35.57 & 20.85 & 41.61 & 31.66 & 31.10 \\
W.V. & \checkmark & 38.01 & 21.99 & 16.23 & \textbf{71.21} & 22.60 & 42.22 & 32.40 & 33.23 & \textbf{40.86} & \textbf{32.82} & 25.05 & 24.25 &  33.40 \\
\textit{one} S.F.  &\checkmark & 50.34 & 46.93 & 51.09 & 49.21 & 26.40 & 39.74 & 39.66 & 36.84 & 33.20 & 26.69 & \textbf{44.50} & 37.92 & 40.21 \\
\textit{five} S.F. & \checkmark & \textbf{58.31} & \textbf{62.78} & \textbf{65.39} & 40.52 & 20.28 & 51.12 & \textbf{40.34} & \textbf{54.35} & 23.37 & 27.47 & 38.82 & 47.52 & \textbf{44.19} \\
one S.F. + W.V.  & \checkmark & 53.64 & 54.01 & 53.84 & 29.74 & 25.42 & \textbf{57.02} & 37.83 & 35.62 & 29.26 & 32.13 & 34.53 & \textbf{49.68} & 41.06 \\
\textit{one} P.F. & \checkmark & 35.26 & 70.65 & 57.35 & 63.34 & 40.64 & 36.44 & 51.68 & 58.94 & 29.20 & 55.17 & 71.98 & 73.39 & 53.67 \\
\textit{five} P.F. & \checkmark & 28.75 & 81.13 & 54.73 & 90.49 & 28.87 & 34.73 & 49.89 & 73.86 & 34.91 & 55.66 & 80.67 & 78.59 & 57.69 \\
\hline
\end{tabular}
}
\caption{\small Photo classification accuracy on Sketchy Dataset: coarse to fine-grained, holding out out 3 coarse categories. Abbreviations as in Table~\ref{category-level binary}. CG indicates coarse category known at runtime or not. Best non-photo results are in bold.}\label{coarse2fine-grained}
\end{center}
\vspace{-1.15em}
\end{table*}

\vspace{-1.4em}
\keypoint{Settings:} Coarse photo models are trained by taking photos from all fine-grained categories in one group as positive examples and the other (non-overlapping with Sketchy dataset) photos from ImageNet as negative examples. Fine-grained photo models are trained by taking photos from one fine-grained category as positive examples.  In this setup the model regressor inputs both a coarse-grained photo classifier and a fine-grained sketch feature\footnote{Since the prior experiments showed sketch feature input was typically better than sketch model input, we stick to sketch feature input here.}, and predicts the corresponding fine-grained photo classifier. 
\doublecheck{For each coarse training category, 500 SVM models are trained by taking 250 photos in $this$ coarse category as positive examples and the same number of random negative samples from the remaining training categories.} \\
In this experiment we train on 17 coarse categories and select three coarse categories for testing. Evaluation uses binary AP. We consider two conditions: (1) Coarse category unknown at runtime. AP  is evaluated among all 12 fine-grained categories in the 3 held out coarse categories. (2) Coarse category assumed known at runtime. We only differentiate among the constituent fine-grained categories in AP evaluation.

\keypoint{Results:} 
From the results in Table~\ref{coarse2fine-grained} we can see that: 
(i) Sketch-based coarse$\to$fine regression does indeed provide a reasonable fine-grained photo classifier that outperforms the NN  + S.F. baseline. (ii) As in the previous experiments, availability of more sketches is still beneficial. (iv) The sketch-based coarse $\to$ fine classifier  clearly outperforms the word-vector baseline (S.F. \vs W.V.) supporting the efficacy of this novel input modality for classifier generation. (v) There is a slight performance increase when combining sketch features and word-vectors.
\vspace{-0.8em}
\subsubsection{Visualization}
\begin{figure}[t]
\centering
\begin{subfigure}[h]{0.97\linewidth}
\includegraphics[width=0.23\linewidth]{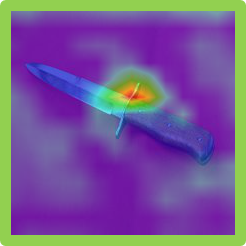}
\includegraphics[width=0.23\linewidth]{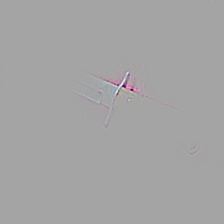}
\hspace{0.2em}
\includegraphics[width=0.23\linewidth]{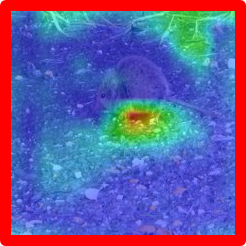}
\includegraphics[width=0.23\linewidth]{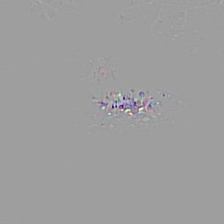}
\\
\includegraphics[width=0.23\linewidth]{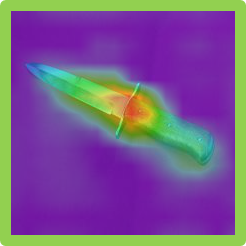}
\includegraphics[width=0.23\linewidth]{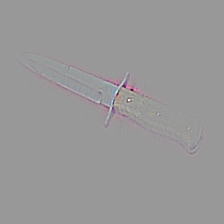}
\hspace{0.2em}
\includegraphics[width=0.23\linewidth]{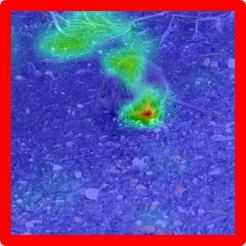}
\includegraphics[width=0.23\linewidth]{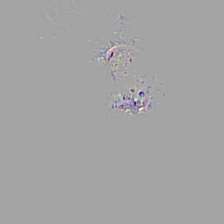}
\end{subfigure}
\\
\vspace{0.1em}
\begin{subfigure}[h]{0.97\linewidth}
\includegraphics[width=0.23\linewidth]{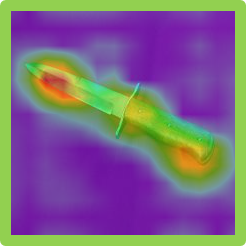}
\includegraphics[width=0.23\linewidth]{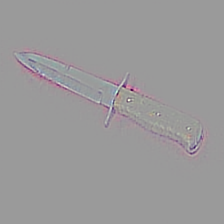}
\hspace{0.2em}
\includegraphics[width=0.23\linewidth]{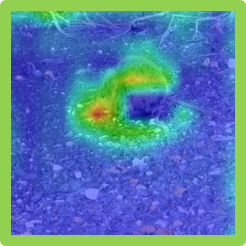}
\includegraphics[width=0.23\linewidth]{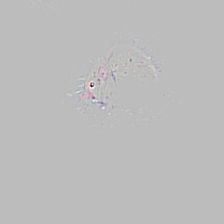}
\\
\includegraphics[width=0.23\linewidth]{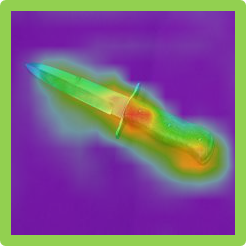}
\includegraphics[width=0.23\linewidth]{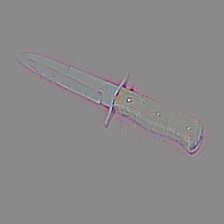}
\hspace{0.2em}
\includegraphics[width=0.23\linewidth]{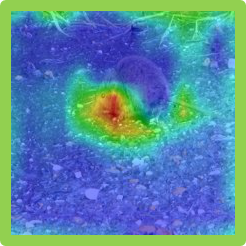}
\includegraphics[width=0.23\linewidth]{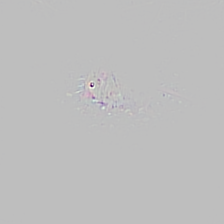}
\end{subfigure}
\caption{\small CNN photo classification decision process visualized by GradCAM \cite{gradcam-iccv17}. Rows: 1. W.V. regressed model. 2. Raw sketch model. 3. F2M Sketch Regressed model. 4. Ground truth many-shot photo model. Left: Region of interest heatmap (images recognized correctly have a green border, and those recognized incorrectly have red). Right: Important pixels for recognition.}
\label{fig:gradcam}
\vspace{-0.3em}
\end{figure}

We finally provide a qualitative visualization of the computation of our sketch-based classifier synthesis compared to alternatives. We exploit the recent deep network analysis tool GradCAM \cite{gradcam-iccv17} to visualize the region of interests of the  classifiers in some example images (Figure~\ref{fig:gradcam}). 
We compare four models: 1. Word-vector regressed models. 2. Directly applied sketch models, raw sketch model. 3. Regressed sketch model. 4. Ground truth photo model. In each case we replace the normal output layer of VGG-19 with the corresponding classifier and apply GradCAM to visualize its reasoning.
From Figure~\ref{fig:gradcam}, we can see some differences: The word-vector regressed model (first row) doesn't look at the full spatial extent of the knife, and it is more distracted by background texture in the case of the mouse. The raw sketch model (second row) gets closer to the extent of the knife but is still distracted by background of the mouse. The regressed sketch model (third row) better estimates the extent of the objects, doing so comparably to the ground truth photo model (fourth row). 

\section{Conclusion}
We proposed the novel concept of sketch-based classifier synthesis that provides an alternative to zero-shot learning when categories are easier to draw than to name. Using a model-regression approach we showed that effective photo classifiers can be synthesized using one or few sketches. The approach is synergistic with traditional zero-shot approach of synthesis based on word-vectors, and can be extended to diverse variants such as using sketch to generate a fine-grained classifier from a coarse-grained classifier. In future work we will apply this idea to synthesizing models for other photo tasks such as segmentation.

{\small
\bibliographystyle{ieee}
\bibliography{egbib}
}

\end{document}